# Smart Magnetic Microrobots Learn to Swim with Deep Reinforcement Learning

## Authors

Michael R. Behrens[1]

Warren C. Ruder*[1,2]

## Affiliations

[1]Department of Bioengineering, University of Pittsburgh; Pittsburgh, PA, USA.

[2]Department of Mechanical Engineering, Carnegie Mellon University; Pittsburgh, PA, USA.

*Corresponding author Email: warrenr@pitt.edu

## One-Sentence Summary

Deep reinforcement learning was used to autonomously develop a robust controller for helical magnetic microrobots.

## Abstract

Swimming microrobots are increasingly developed with complex materials and dynamic shapes and are expected to operate in complex environments in which the system dynamics are difficult to model and positional control of the microrobot is not straightforward to achieve. Deep reinforcement learning is a promising method of autonomously developing robust controllers for creating smart microrobots, which can adapt their behavior to operate in uncharacterized environments without the need to model the system dynamics. Here, we report the development of a smart helical magnetic hydrogel microrobot that used the soft actor critic reinforcement learning algorithm to autonomously derive a control policy which allowed the microrobot to swim through an uncharacterized biomimetic fluidic environment under control of a time varying magnetic field generated from a three-axis array of electromagnets. The reinforcement learning agent learned successful control policies with fewer than 100,000 training steps, demonstrating sample efficiency for fast learning. We also demonstrate that we can fine tune the control policies learned by the reinforcement learning agent by fitting mathematical functions to the learned policy's action distribution via regression. Deep reinforcement learning applied to microrobot control is likely to significantly expand the capabilities of the next generation of microrobots.

## Introduction

Untethered swimming microrobotic systems have received significant research attention for performing micromanipulation tasks and particularly for their potential therapeutic biomedical applications [1, 2]. Microrobots operating remotely inside the human body have potential to enable minimally invasive medical procedures including targeted drug, cell, or other cargo delivery [3-6], tissue biopsy [7], thermotherapy [8], and blood clot removal [9]. Controlling these miniature devices within complex and dynamic environments such as the human body can present a significant engineering challenge [10]. This challenge is in part because the design of microrobotic systems is trending towards the use of complex composite materials [6, 11-13], dynamic morphologies [14-17], and integrated biological components [18-22]. These features add layers of functionality to microrobotic systems, but can create difficulties when constructing accurate dynamic and kinematic models of microrobotic behavior, making it especially complex and challenging to use classical feedback control systems to control microrobot behavior [5, 14, 23]. Additionally, the environmental dynamics encountered by a biomedical microrobot inside the human body may be variable, complex, and poorly characterized [16, 24, 25].

As one potential pathway to overcome these challenges, we can observe and adopt the strategies of natural biological agents that have evolved to operate in complex, unpredictable environments. Many biological systems can adapt to learn new behaviors based on experience, allowing them to thrive in a wide range of complex and variable environmental conditions by tailoring their behavior to suit the environment. Systems capable of learning adaptive behavioral patterns based on past events are ubiquitous in nature and are found across all levels of biological hierarchy, including in biochemical networks [26], bacteria [26, 27], nematode worms [28], insects [29], plants [30], adaptive immune systems [31], and animal behavior [32]. Inspired by the wide-ranging applicability of adaption and learning to the success of living organisms, engineered microrobotic systems that learn new behaviors from past experience could enable new capabilities for complex microrobotic systems [33].

Reinforcement learning (RL) is a biomimetic machine learning optimization technique inspired by the adaptive behavior of real-world organisms [34] that can enable learning behaviors in artificial engineered systems [35]. In RL, an agent observes the state of an environment, and chooses actions to perform in the environment to achieve a task specified by a reward signal, which is typically predefined. The reward signal is used to teach the agent to perform actions to maximize the expected future rewards, which enables the agent to learn to perform the task better based on past experience. RL algorithms have achieved success in a range of complex robotic control applications [35-40]. For example, RL for robotic control has been demonstrated to create robotic control policies that achieve better performance than many humans at complex tasks such as grasping and accurate throwing of irregularly shaped objects into bins [39]. RL algorithms have also been shown to exceed human level performance in complex virtual tasks with large possible state spaces that cannot be tractably and exhaustively modeled, such as the game of Go [41]. Machine learning techniques including RL have already demonstrated promise for developing policies to control microrobot behavior. In simulation, RL agents have been trained to control microrobot behavior for solving navigation and swimming challenges in heterogenous fluids [42, 43]. An early RL algorithm, Q-learning, has also been shown to be effective for controlling the behavior of laser-driven microparticles in a discretized grid environment [44]. Other similar machine learning techniques such as Bayesian optimization have been demonstrated to learn walking gates for difficult-to-model magneto-elastomeric millirobots [45]. However, control of swimming microrobots that use deep reinforcement learning to operate in dynamic, biomimetic, and microfluidic environments with clinically relevant magnetic actuation has yet to be reported.

In this work, we demonstrate that deep reinforcement learning based on the Soft Actor Critic algorithm [46] can be used to create smart soft helical magnetic microrobots that autonomously learn optimized swimming behaviors when actuated with non-uniform, nonlinear, and time-varying magnetic fields in a physical fluid environment. Our RL microrobots learned successful actuation policies without any *a priori* knowledge about the dynamics of the microrobot, the electromagnetic actuator, or the environment (Figure 1a). These results demonstrate the potential of reinforcement learning for developing high performance multi-input, multi-output (MIMO) controllers for microrobots without the need for explicit system modeling. This capability to autonomously learn model-free microrobot control algorithms could significantly reduce the time and resources required to develop high performance microrobotic systems [33].

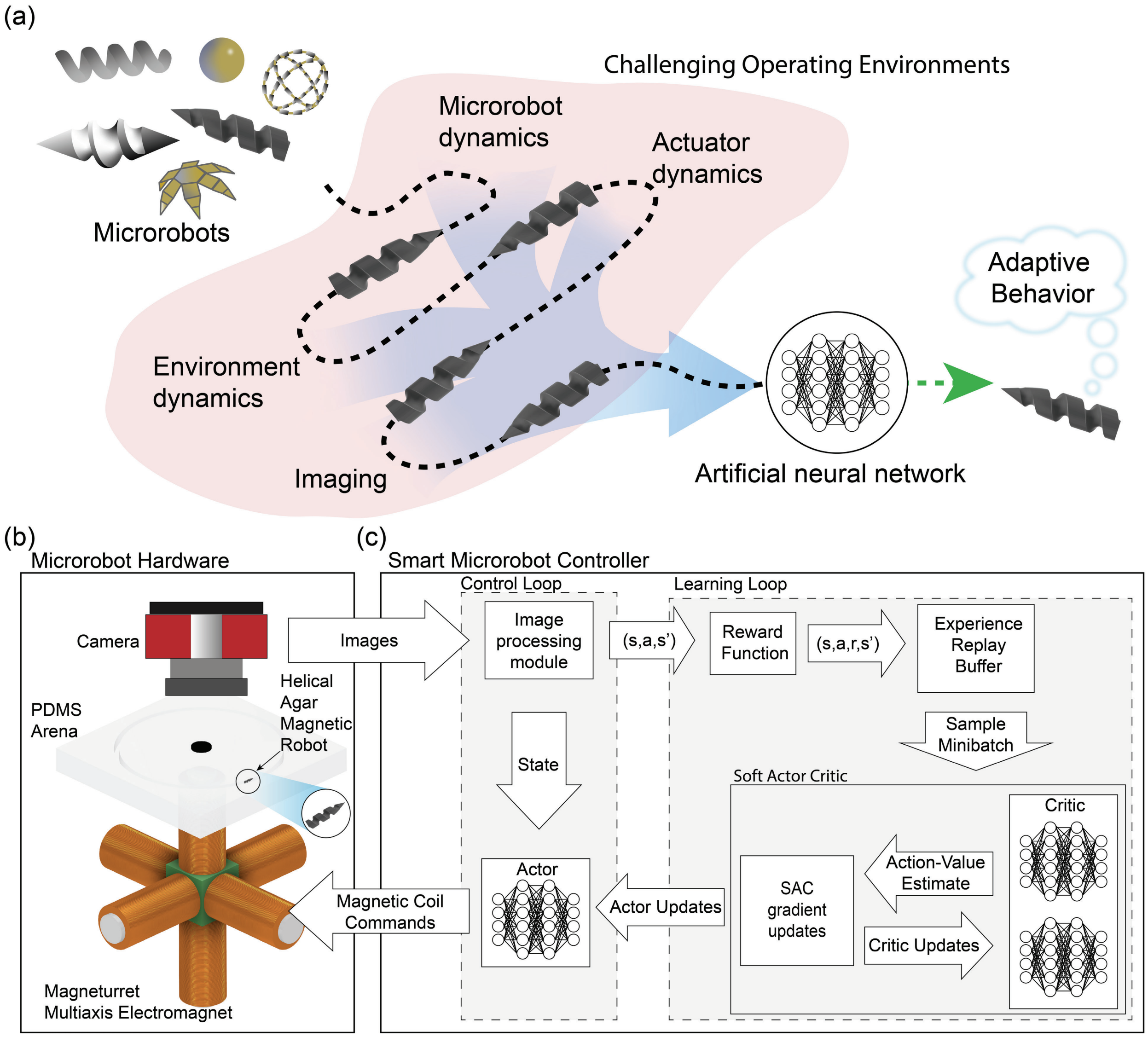

**Figure 1. Microrobots with unknown dynamics in uncharacterized environments can be controlled with deep reinforcement learning. (a)** Microrobotic systems are designed with a great variety of shapes, sizes, materials, and actuation methods, and are often operated in challenging environments. Controllers based on artificial deep neural networks trained with reinforcement learning (RL) can factor in all of these complex dynamic systems and inputs to create model-free microrobot controllers to create adaptive microrobots. **(b)** Our microrobotic system consisted of a helical agar magnetic robot (HAMR) in a circular polydimethylsiloxane (PDMS) fluidic track that was given the task of moving to a target position along the track under control of a multi-axis electromagnet (Magneturret). Images of the HAMR in the arena were captured with an overhead camera. **(c)** The smart microrobot control system used a neural network trained with the Soft Actor Critic (SAC) reinforcement learning algorithm to generate commands for the Magneturret. In the control loop, the stream of images from the overhead camera was processed to generate state information that was then fed into the actor neural network, which returned a set of continuous actions that were used to control the currents in the Magneturret. State (s), action (a), reward (r), and next state (s') information was stored in a replay buffer which was used update actor and critic neural networks off policy in a learning loop.

## Results

In order to create an environment where we could test the hypothesized efficacy of RL control systems for microrobots, we first designed and built a physical, biomimetic, fluidic arena with multidimensional magnetic actuation, and deployed a magnetic microrobot in the arena whose design was inspired by the work of Kumar and colleagues [4]. In our experimental setup (Figure 1b), a helical agar magnetic robot (HAMR) was tasked with swimming clockwise through a fluid filled lumen in a polydimethylsiloxane (PDMS) arena under control of a non-uniform rotating magnetic field generated by a three-axis array of electromagnetic coils (Magneturret). An overhead camera was used to track the position of the HAMR in the channel. The camera was used to pass images to a control algorithm consisting of an image processing module and a neural network which generated commands for the Magneturret (Figure 1c). The goal of the control system for our remotely actuated microrobot was to manipulate the shape and magnitude of the actuating energy field in order to move the microrobot to achieve an intended dynamic behavior. The controller neural network was trained via a reinforcement learning agent using the soft actor critic algorithm.

The fundamental control problem was encapsulated by this question: how should the currents in the electromagnetic coils be modulated in order to create a magnetic field that places forces and torques on the HAMR sufficient to drive its locomotion toward a specific target? Achieving this task usually requires an accurate dynamic model of the complete system, including the dynamics of the robot, the environment, and the actuator [23]. Significant work has been done developing dynamic and kinematic models for different microrobots and actuators [14, 47-49]. These models are often developed by making simplifying assumptions about the system such as uniform magnetization [47], ideal shape [47], and system linearity [50, 51], which could lead to behavioral deviations between the physical system and the modeled system. The difficulty in accurately modeling the dynamics of microrobot behavior increases significantly for microrobots with complex magnetization profiles, soft material composition, or active shape-changing capabilities [13, 14, 17, 52, 53].

Instead of explicitly modeling the dynamics of the magnetic actuator and the HAMR within the environment and specifying a controller, we performed the much simpler task of specifying the desired behavior of the HAMR in the form of a reward signal. The agent observed the state of the environment along with a reward signal containing information about actions that lead towards the successful completion of the task. The RL agent started without any *a priori* information about the task and had to learn to perform the task by sampling actions from the space of all possible actions and learning which actions resulted in behavior that was rewarded.

The RL controller required us to develop and formulate the task as well as the associated reward signal. At the beginning of each training episode, a target position was defined, 20° clockwise from the starting position of the HAMR in the circular channel. The objective of the RL agent was to develop an action policy, $\pi$, which maximized the total value of the rewards it would receive if it followed that policy in the future. When the environment was in state, $s$, the agent chose an action, $a$, from the policy according to $a \sim \pi(\cdot\, |s)$, probabilistically selecting from a distribution of possible actions available in that state. The agent received a reward when it selected actions that moved the HAMR clockwise through the circular lumen towards the target, and it received a negative reward when it moved the HAMR counterclockwise. If the HAMR reached the target within the allotted time, the agent was given a large bonus reward, and the target position was advanced 20°. The reward function we selected was

$$r(s,a) = \Delta\theta_r + 1000 \textbf{ if } (\theta_r = \theta_g) \qquad (1)$$

where $\theta_r$ is the angular position of the HAMR in the channel in degrees, $\theta_g$ is the angular position of the goal, and $\Delta\theta_r$ is the change in angular position of the HAMR as a result taking of action $a$ in state $s$.

The reinforcement learning problem was formalized as a Markov decision process [34], in which at time $t$, the state of the system $s_t$ is observed by the agent. The agent performs an action $a_t$ which changes the state of the environment to $s_t{}'$, yielding a reward $r_t(s_t, a_t)$. This process continues for the duration of the task, yielding a trajectory of the form $(s_t, a_t, r_t, s_{t+1}, a_{t+1}, r_{t+1}, s_{t+2} \ldots)$. The goal of the RL agent is to identify an optimal policy $\pi^*(a|s)$ for selecting actions, based on state observations, that maximize the rewards received for following the policy. Over the course of training, the agent autonomously learned a control policy by trying actions in the environment, observing the reward obtained by performing those actions, and modifying its future behavior in order to maximize the expected future return.

The control problem for our microrobotic system was formulated as an episodic, discrete time problem with a continuous action space and continuous state space. The state space consisted of all the possible states for the system: the position of the HAMR within the channel, the speed of the HAMR, the shape of the magnetic fields, the time remaining in the episode, and relative position of the robot to the target position in the channel. The action space consisted of four continuous actions, which controlled the magnitudes and phase angles for sinusoidal currents in the Magneturret. While the current waveforms could theoretically take on an infinite number of shapes, we chose to define the applied waveforms as sinusoids to bound the space of possible actions that the agent could take. Sinusoidal currents were chosen because these can be used to generate rotating magnetic fields in other three-axis electromagnetic actuators for microrobots, such as Helmholtz coils [54].

## Entropy regularized deep reinforcement learning enabled continuous microrobot control

We selected the Soft Actor Critic RL algorithm (SAC) for this research. SAC is a maximum entropy RL algorithm that seeks to balance the expected future rewards with the information entropy of the policy [46]. In other words, SAC learns a policy that successfully completes the task while acting as randomly as possible, which in practice often leads to robust policies that are tolerant of perturbations in environmental conditions [36]. SAC had previously proven useful for real-world robotic tasks with high-dimensional, continuous state and action spaces [36, 38], which suggested that it would be applicable our microrobotic control problem. In previously reported applications of real-world reinforcement learning with physical systems [55], SAC was demonstrated to be highly sample efficient, requiring relatively few environmental interactions in order to develop a successful policy. Sample efficiency is critical when performing reinforcement learning with real-world robotics (i.e. not simulated) in order to reduce wear and tear on the system, and in order to minimize the time needed to learn a policy [37].

The SAC algorithm seeks to develop an optimal stochastic policy $\pi^*$:

$$\pi^* = arg \max_{\pi} \sum_t \mathbb{E}_{(s_t,a_t)\sim\pi}[r(s_t,a_t) + \alpha\mathcal{H}(\pi(\cdot\,|s_t)] \quad (2)$$

where $\mathcal{H}$ is the information entropy of the policy and $\alpha$ is a temperature hyperparameter, which balances the relative impact of the policy entropy against the expected future rewards. Here, we used a version of the SAC algorithm in which the temperature is automatically tuned via gradient descent so that the entropy of the policy continually matches a target entropy, $\overline{\mathcal{H}}$, which we selected to be -4 (-$Dim$ of the actions space) using a heuristic suggested by Haarnoja et al [56]. A full derivation of the soft actor critic algorithm is beyond the scope of this paper, but interested readers are directed to Haarnoja et al. [46]. Briefly, the SAC algorithm uses an agent called the actor, denoted as $\pi$, which is a deep neural network that takes the state of the system $s_t$ as input, and returns action $a_t$ as output. A value function is created to rate the value of taking actions when in particular states, and instantiated using two critic neural networks $Q_{1,2}(s,a)$ which take states and actions as input, and return values corresponding to the relative value of taking action $a_t$ in state $s_t$. Two Q networks are trained in order to reduce overestimation in the value function. Environmental transitions in the form of $(s,a,r,s',d)$ sets are recorded in an experience replay buffer, $D$, where $d$ is a done flag denoting a terminal state, set either when the microrobot has reached the goal, or when the episode has timed out. The SAC algorithm learns off-policy by randomly sampling minibatches of past experiences from $D$, and performing stochastic gradient descent over the minibatch in order to minimize loss functions for the actor network, $\pi$, critic networks, $Q_1$ and $Q_2$, and temperature parameter, $\alpha$. Over the course of learning, the parameters of the actor and critic neural networks are updated so that the behavior of the policy approaches the optimum policy, $\pi^*$. A detailed version of the soft actor critic algorithm for microrobot control that we used in this study is available in Algorithm 1. Neural network architectures and hyperparameters used are available in Supplementary Table 1.

## Hardware to control magnetic microrobots with reinforcement learning

Magnetic fields created by electromagnetic coils are common actuators used for magnetic microrobots and have significant potential for clinical medical applications [3, 57, 58]. Magnetic fields act on a magnetic microrobot by imparting forces and torques on the robot. For a microrobot with a magnetic moment, $\boldsymbol{m}$, in a magnetic field, $\boldsymbol{B}$, the robot experiences a force $\boldsymbol{F}$ according to

$$\boldsymbol{F} = \nabla(\boldsymbol{m}\cdot\boldsymbol{B}) \quad (3)$$

In a non-uniform magnetic field (i.e., a magnetic field with a spatial gradient), a ferromagnetic or paramagnetic microrobot feels force in the direction of increasing magnetic field gradient. The magnetic microrobot also experiences a torque according to

$$\boldsymbol{\tau} = \boldsymbol{m}\times\boldsymbol{B} \quad (4)$$

which acts to align the magnetic moment of the microrobot with the direction of the magnetic field. When the magnetic field is rotated so that the direction of $\boldsymbol{B}$ is constantly changing, it is possible to use this torque to impart spin to the microrobot at the frequency of the rotating magnetic field, up to the step out frequency of the robot [47]. If the spinning microrobot is helically shaped, rotation can be transduced into forward motion so that the microrobot swims similar to how bacterial are propelled by flagella [59]. This non-reciprocal helical swimming is efficient in low Reynolds number fluidic environments commonly encountered by microrobots [59]. Because of the efficiency of this swimming mode, and because the magnetic torque available to a microrobot decreases more slowly with distance compared to the force [23], magnetic microrobots designed for long range magnetic operation are often helically shaped [6, 11, 60]. For this reason, we selected a helical magnetic microrobot, the HAMR, as our model system.

**Algorithm 1:** Soft Actor Critic for Microrobot Control

Initialize policy parameters $\sigma$, Q-function parameters $\omega_1, \omega_2$, and empty FIFO replay buffer $D$
Set target Q-function parameters equal to main parameters $\omega_{target,n} \leftarrow \omega_n$
Initialize $\bar{\mathcal{H}} = -$ number of actions (4), $\alpha = 1$
Observe initial state $s_{t=0}$, and calculate $\theta_{robot} \in (0,360°)$
Set $\theta_{goal} \leftarrow \theta_{robot} + 20°$, $\theta_{goal} \in (0,360°)$
Data Collection Process: **Repeat**
**If** new $\pi_\sigma$ is available: update
  **While** t steps < 33 **or** done = false
    select action $a_t \sim \pi_\sigma(\cdot\,|s_t)$
    Execute $a_t$ in the environment
    **For** j in range (3)
      Wait 0.3 seconds
      Observe next state $s'_j$, reward $r_j(s'_j)$, and done $d$ (1 **If** $\theta_{robot} = \theta_{goal}$ $Or\ t = 33\ Else$ 0)
    End **For**
    Set $s'_t \leftarrow (s'_{j=1}, s'_{j=2}, s'_{j=3})$, $r_t \leftarrow \sum_j r_j$
    Store transition $(s_t, a_t, r_t, s'_t, d)$ in replay buffer $D$
    Set $s_t = s'_t$
    **If** done: set $\theta_{goal} \leftarrow \theta_{robot} + 20°$,
    t++
  End **While**
Set $\theta_{goal} \leftarrow \theta_{robot} + 20°$, t = 0

Training Process: **Repeat**
**If** number of updates < number of transitions in $D$
  Randomly sample batch of transitions, $B = \{(s, a, r, s', d)\}$ from $D$
  Compute targets for the Q functions:
$$y(r, s', d) = r + \gamma(1-d)\left(Q_{\omega_{target,i}}(s', \tilde{a}') - \alpha\, log\, log\, \pi_\sigma(s')\right), \tilde{a}' \sim \pi_\sigma(\cdot\,|s')$$
  Update Q-functions using
$$\nabla_{\omega_i} \frac{1}{|B|} \sum_{(s,a,r,s',d)\in B} \left(Q_{\omega_i}(s,a) - y(r,s',d)\right)^2 \qquad \text{for } i = 1,2$$
  Update Policy using
$$\nabla_\sigma \frac{1}{|B|} \sum_{s\,\in B} \left(Q_{\omega_i}(s, \tilde{a}_\sigma(s)) - \alpha\, log\, log\, \pi_\sigma(\tilde{a}_\sigma(s)|s)\right),\ \tilde{a}_\sigma \sim \pi_\sigma(\cdot\,|s)$$
  Update temperature $\alpha$ using
$$\nabla_\sigma \frac{1}{|B|} \sum_{s\,\in B} \left(-\alpha\, log\, log\, \pi_\sigma(a|s) - \alpha \underline{H}\right),\ a \sim \pi_\sigma(\cdot\,|s)$$
  Update the target Q-functions using
$$\omega_{target,n} \leftarrow \omega_{target,n} + (1-\tau)\omega_n \qquad \text{for } n = 1,2$$
End **If**
Send latest $\pi_\sigma$ to data collection process every minute
**Until** convergence

The HAMR that we created for this study was composed of a 2% w/v agar hydrogel, which was uniformly diffused with 10% w/v iron oxide nanopowder to form a magnetically susceptible soft polymer (Figure 2a) [4]. This magnetic agar solution was heated to melting temperature and a syringe was used to inject the liquid into a helical mold created using a stereolithography 3D printer (Figure 2b). The agar in the mold solidified and the robots were removed with a metal needle and stored long-term in deionized (DI) water. The HAMRs molded for this study were 4.4 mm in length, 1 mm in diameter, and asymmetrical from head to tail, with flat head and a pointed tail (Figure 2c,d). Microrobots formed with this technique have been previously shown to be controllable within rotating magnetic fields, and to perform biomedical functions such as cell delivery [4] and active biofilm removal in the root canal of human teeth [54]. For our application, this HAMR design had several advantages. The HAMRs were simple to manufacture at low cost with batch fabrication methods. The HAMRs were small enough to act as helical swimming robots in a flow regime with Reynolds number ~1, but large enough, about the size of a small grain of rice, to be easily manipulated and visualized without the use of microscopes or other micromanipulation tools. Because the HAMRs swim with non-reciprocal, helical motion in the presence of a rotating magnetic field, a very common motif in microrobotic research [11, 60, 61], insights gained from this study could readily be extended to other microrobotic systems with similar characteristics. Because the HAMRs were made of soft hydrogel, they were flexible and deformable. Soft bodied robots have many favorable characteristics for *in vivo* use such as deformability to fit through irregular shaped channels and enhanced biocompatibility (e.g., by matching the elastic modulus of the robot to the biological environment) [62]. These characteristics make soft-bodied microrobots appealing for biomedical applications, but it can be more difficult to create accurate dynamic models for soft-bodied microrobots [16]. Our method of using reinforcement learning to develop control systems without explicit modeling could be particularly useful for soft microrobots due to this modeling constraint. Finally, despite being soft-bodied, the hydrogel structure of the HAMR did not experience noticeable wear over the course of several months of continuous use, thus meeting a practical reinforcement learning constraint that the system not be susceptible to significant wear and tear during extended use which would cause a distribution shift in the collected data as the dynamic properties of the system degraded [55].

As an actuator for our microrobot system, we developed a three-axis magnetic coil actuator –the Magneturret– which contained six permalloy-core magnetic coils arranged on the faces of a 3D-printed acrylonitrile butadiene styrene (ABS) plastic cube (Figure 2e). The two coils on opposite sides of the central cube along each axis were wired together in series so that they both contribute to the generation of a magnetic field along their respective axis. Each of the three coils, hereafter referred to as the X, Y, and Z coils, were driven with a sinusoidal current generated by a pulse width modulated (PWM) signal created by a microcontroller and amplified in an H-bridge motor driver. The resulting magnetic field, produced by the superposition of the magnetic fields from the three coils, could be modulated by varying the frequency, amplitude, and phase angle of the sine current waves in each coil. To cool the coils and prevent thermal damage, the Magneturret was sealed with epoxy resin into a 3D-printed housing and coolant was continuously pumped through while the coil was operating (Figure 2f). The RL agent was given direct control over the magnitude and phase angles of the sinusoidal driving currents in the X and Y-axis coils of the Magneturret (Table 1). The Z-axis magnitude was calculated as the larger of the two magnitudes in X and Y, and the Z-axis phase angle was fixed. The sinusoidal currents in each axis used a fixed angular frequency of 100 rad/s (15.9 Hz).

***Magnetic Coil Control Parameters***

| **Control Variable** | **Symbol** | **Source** | **Range** |
|---|---|---|---|
| Frequency | $f$ | Fixed | 15.9 Hz ($\omega=2\pi f=100$ rad/s) |
| Magnitude X | $M_X$ | RL agent | [-1,1] unitless |
| Magnitude Y | $M_Y$ | RL agent | [-1,1] unitless |
| Magnitude Z | $M_Z$ | $\max(\lvert M_X\rvert,\lvert M_Y\rvert)$ | [0,1] unitless |
| Phase Angle X | $\varphi_X$ | RL agent | [0,2π] radians |
| Phase Angle Y | $\varphi_Y$ | RL agent | [0,2π] radians |

***Control Equations***

| | | |
|---|---|---|
| Current in X-axis coil | $I_x = M_x \sin(ft + \varphi_x)$ | (5) |
| Current in Y-axis coil | $I_y = M_y \sin(ft + \varphi_y)$ | (6) |
| Current in Z-axis coil | $I_z = M_z \sin(ft)$ | (7) |

**Table 1. Control inputs for electromagnet waveforms.**

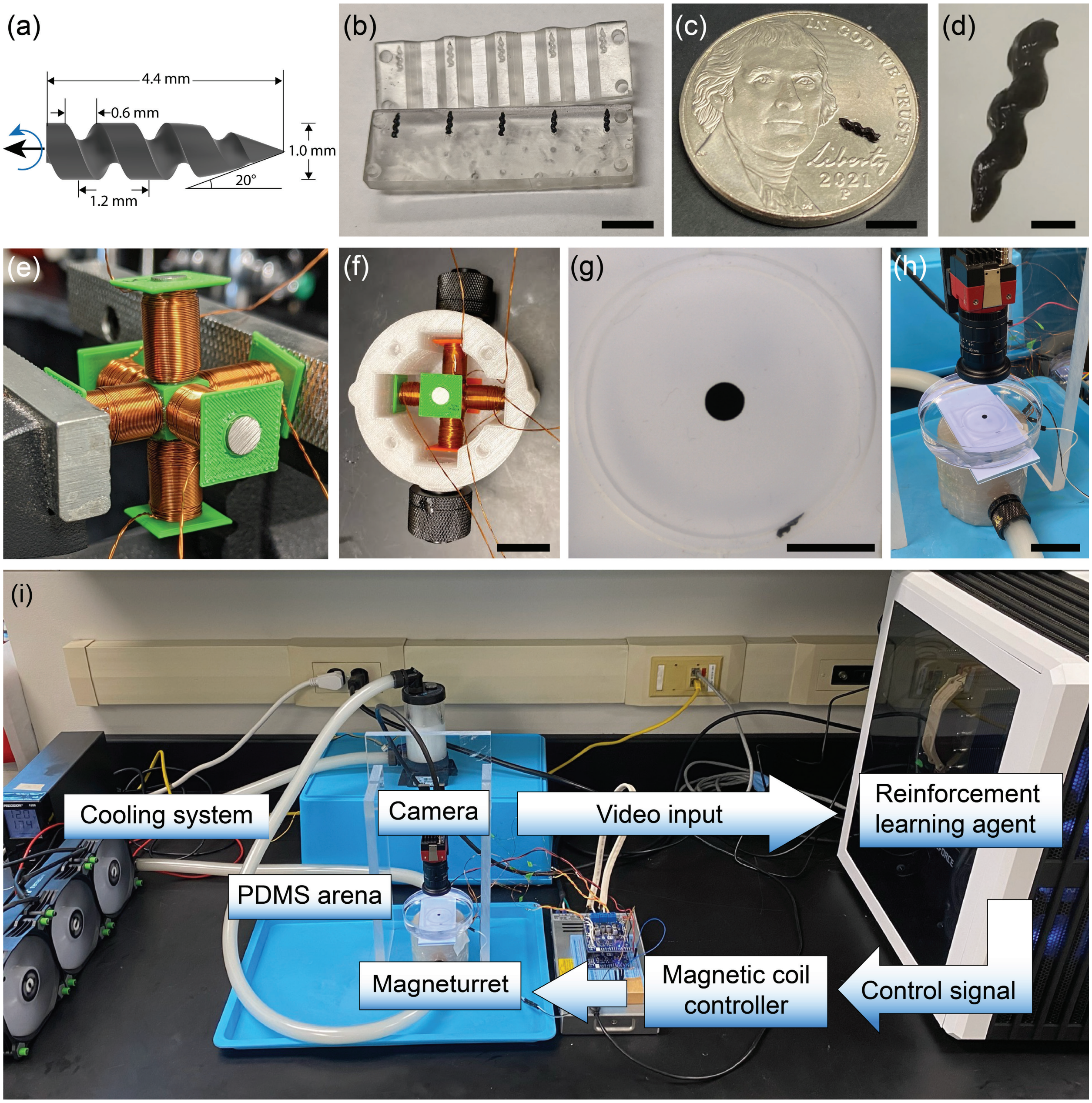

**Figure 2. Hardware for real-world control of magnetic microrobots using reinforcement learning. (a)** Helical Agar Magnetic Robot (HAMR) schematic. **(b)** HAMRs were fabricated by molding molten magnetic hydrogel in 3D printed molds. Scale bar = 10 mm. **(c)** HAMR with a United States 5-cent coin. Scale Bar = 5 mm. **(d)** HAMRs were composed of agar hydrogel infused with iron oxide nanopowder. Scale bar = 1 mm. **(e)** The Magneturret was composed of six coils of copper magnet wire wrapped around permalloy cores, positioned on the faces of a central 3D printed cube. **(f)** The Magneturret was enclosed within a 3D printed housing and sealed with epoxy, and glycerol coolant was continuously pumped through the Magneturret housing. Scale bar = 20 mm. **(g)** A circular PDMS track with a rectangular cross section used as an arena for the HAMR. A black, circular fiducial marker indicates the center of the arena. Scale bar = 10 mm. **(h)** The PDMS arena was submerged in a water-filled petri dish placed on top of the Magneturret, with an acrylic LED light sheet as a backlight for uniform bottom-up illumination. Scale bar = 30 mm. **(i)** The complete hardware system.

We chose to operate our HAMR in a circular, fluid-filled track for this study. This arena served as a simple environment which mimics the tortuous *in vivo* luminal environments that microrobots operating in the body might encounter, while providing a simple environment for us to establish a robust proof-of-concept reinforcement learning control system (Figure 2g). The HAMR could swim in a complete circle within this arena, and no human intervention was required to reset the position of the robot in the environment during training, which facilitated automated learning [37]. The arena was constructed by pouring polydimethylsiloxane (PDMS) over a polyvinyl chloride ring with an outer diameter of 34 mm and a 1.7 mm x 3 mm rectangular cross section. The PDMS was then cured, and plasma bonded to a second flat sheet of cured PDMS to form a rectangular lumen for the HAMR to swim. PDMS is transparent, allowing us to see the robot in the arena and to visually track it with an overhead camera. During long-term learning experiments, the PDMS arena was submerged in a petri dish filled with DI water in order to prevent the formation of air bubbles in the channel due to evaporation over the course of an experiment. This petri dish was then placed on top of the Magneturret, with the center of the Z-axis coil aligned with the center of the circular track (Figure 2h). A black rubber wafer was placed into the center of the arena on top of the PDMS to act as a fiducial marker so that the center of the arena could easily be identified with image processing. A diffuse white LED backlight was positioned between the Magneturret and the PDMS arena for uniform bottom-up illumination which facilitated simple image processing by binary thresholding to identify the position of the microrobot in the channel.

We did not perform an extensive analysis to identify the shape or magnitude of the magnetic field created by the Magneturret, or to model the swimming dynamics of the HAMR in the PDMS arena. We hypothesized that we would be able to develop a high-performance control system using RL without going through the effort of developing a system model first.

**Reinforcement learning can be used to learn microrobot control policies**

State information of the microrobotic system was derived by using image processing to create a state vector-based input, which was passed to the RL agent. The angular position, $\theta_r$, of the microrobot in the channel was calculated with image processing by binary thresholding and simple morphological operations. The camera was deliberately run with a slow shutter speed so that the images were intentionally washed out to remove noise. This simplified the task of using binary thresholding operations to identify the position of the HAMR and the center of the channel. The angular position of the HAMR in the channel was measured relative to the fiducial marker in the center of the circular arena. This information, as well as the position of the goal, $\theta_g$, the last action taken by the agent $(M_{x,t-1}, M_{y,t-1}, \varphi_{x,t-1}, \varphi_{y,t-1})$, and the time, $t$, remaining in the episode were used to create a state vector.

Reinforcement learning is based on the mathematics of Markov decision processes, which theoretically require the full state of the system to be available to the agent in order for convergence to be guaranteed [34]. In our particular implementation, the velocity of the HAMR at any given time could not be determined from a single still-frame observation, so the total state of the system given to the agent at each time step was composed of three concatenated sub-observations taken 0.3 seconds apart, for a total step time of 0.9 seconds (see Algorithm 1, steps 11-15). This allowed the agent to infer the velocity of the HAMR based on differences between the three sub-observations. This technique of batching sequential observations for improving the observability of the system for RL has been used successfully in domains such as Atari video games, in which the agent learned from raw pixel data gathered from sequential screenshots of the game [63].

At the beginning of each learning trial, the actor and critic neural network parameters were randomly initialized. We allowed the RL agent to train for a maximum of 100,000 times steps, using a fixed ratio of one gradient update per environmental step, which has been shown to reduce training speed in exchange for higher training stability [64]. 100,000 environment steps were adequate time for effective actuation policies to be learned that continuously moved the HAMR clockwise around the arena (Figure 3a). It is commonly reported when using reinforcement learning that several million environmental steps are necessary to derive a successful policy [37, 63], so this result shows the sample efficiency of SAC, which is critically important for RL tasks that are trained using physical systems instead of in simulation. A time-lapse movie of the HAMR recorded during the learning process is shown in supplementary movie 1. We tracked the net movement of the HAMR during the training process, and each training session ended with the microrobot going continuously around the track in a clockwise direction (Figure 3b).

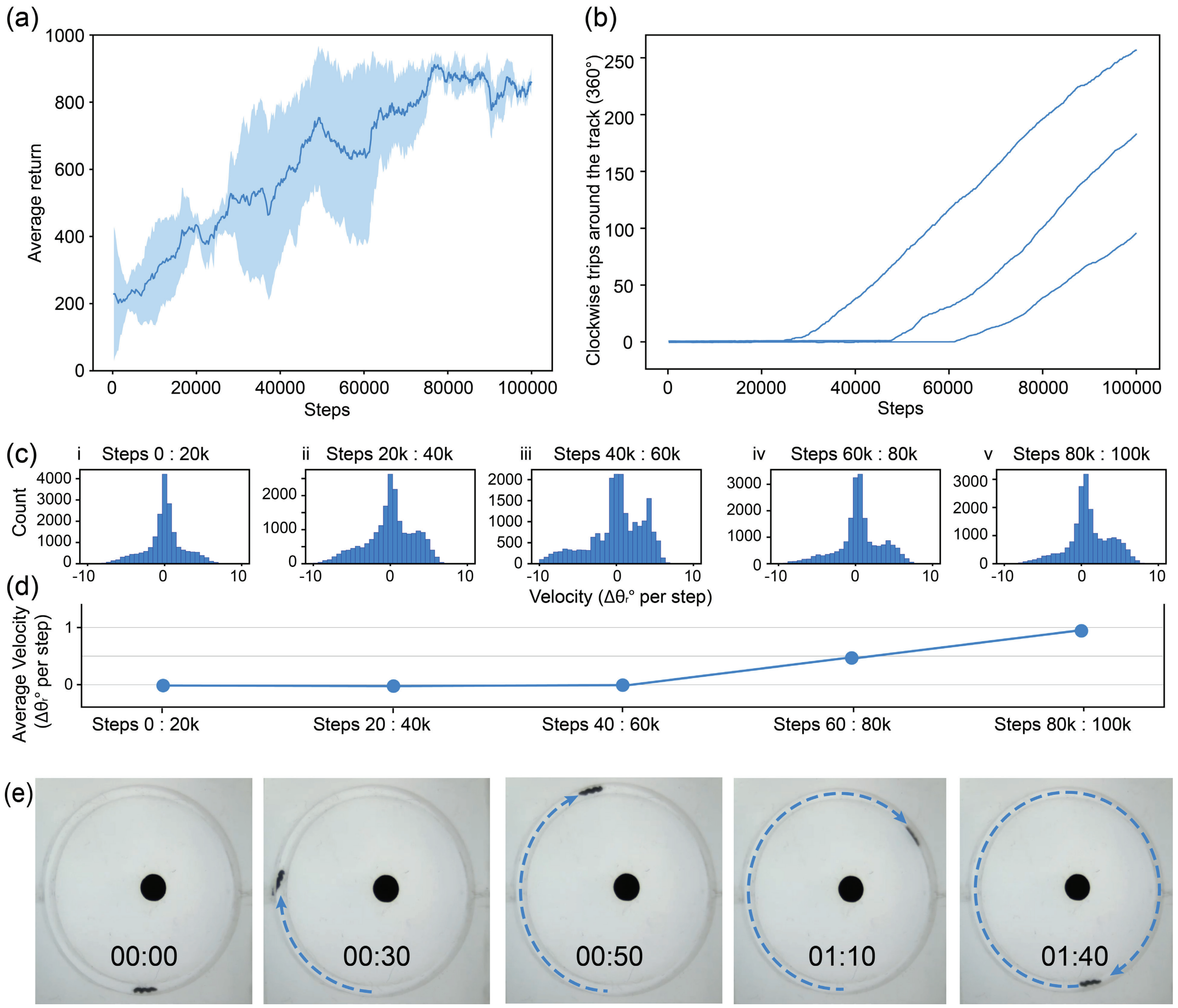

**Figure 3. Reinforcement learning yielded successful control policies for the HAMR within 100k time steps. (a)** We trained the agent for a total of 100k time steps per training session. The trace represents the average and standard deviation of the return from three successful training runs. **(b)** Successful training resulted in policies which, after an initial learning period, achieved consistent forward motion, with the HAMR moving continuously around the circular arena. **(c)** We recorded the actions and state of the agent over 100k training steps. The majority of actions taken by the agent for the first 20k steps resulted in no motion (i). As the agent learned, the action distribution became bimodal (ii-v), with a second peak appearing at ~5 degrees per action, indicating the agent was increasingly taking actions which resulted in forward movement. **(d)** Averaging the velocity distribution over the training period showed that at some time between 40k and 60k steps the agent had learned to achieve net positive movement. **(e)** The RL agent learned policies that moved the microrobot around the full circular arena with helical swimming motion.

We recorded each action taken by the agent during training sessions and the resultant change in state of the microrobot. For a single training run, we plotted the distribution of actions as a function of the resultant change in HAMR position, $\Delta\theta_r$ (Figure 3c). We separated the total 100k steps into 5 bins of 20k steps each. The distribution of actions over the first 20k time steps (Figure 3c, i) is centered around a sharp peak of actions which result in no net movement, as we would expect from an agent with little experience randomly exploring the space of possible actions. By the second batch of 20k steps (Figure 3c, ii), a pattern emerged in which the action distribution shifted to a bimodal distribution in which most actions still result in no net movement, but a second peak on the positive side indicates a trend towards selecting actions which result in clockwise movement. However, during this phase of training, the net motion of the robot remained close to zero (Figure 3d), because of fattening of the negative tail in the action distribution. As the learning process continued, the distribution continued to shift until the average movement was clockwise, with a second peak around 5 degrees per time step, and a narrow tail representing few actions, which caused the robot to move in the counterclockwise direction (Figure 3c, v).

Once the training sessions were complete, we evaluated the learned policies to test their performance. For evaluating policies, we used the highest performing policy parameters learned during a training session by monitoring a rolling average of the return over the last 100 episodes and saving the policy parameters each time the rolling average performance exceeded the last best performing model (Supplementary Figure 1). This was done because we sometimes observed a drop in performance after the peak performance was achieved in training, possibly due to overfitting. Early stopping, or selecting a policy before performance degradation has occurred, is a common technique used to prevent overfitting in neural networks [65]. Successful policies were able to move the robot indefinitely around the complete circular track (Figure 3e, Supplementary Movie 2).

The soft actor critic algorithm learns a continuous stochastic policy, $\pi$, sampling actions from the policy according to $a_t \sim \pi(\cdot\,|s_t)$, in which the actions selected during training are randomly sampled from a Gaussian distribution, and the agent learns the mean μ and the variance of this distribution over the course of training [46]. This is done in order to explore the space of possible actions during training. During training the agent seeks to balance the sum of future rewards with the information entropy of the policy by maximizing an entropy regularized objective function, and the policy entropy corresponds to the explore/exploit tradeoff the agent makes during training. However, once the policies were trained, performance during policy evaluation could be increased by selecting actions from the mean of the distribution without further stochastic exploration according to $a_t = \mu(s_t)$. This deterministic evaluation led to an increase in the proportion of actions taken by the agent which resulted in positive motion (Figure 4a). We compared the total average velocity achieved by all the trained policies in both deterministic and stochastic action selection modes, which showed that deterministic action selection led to higher performance (Figure 4b).

We next examined the distribution of the action values chosen by the RL agent when evaluated deterministically according to $a_t = \mu(s_t)$. For each of the four actions ($M_x, M_y, \varphi_x, \varphi_y$) (Figure 4c) taken by the policy over 3000 time steps, we plotted the value of the action against the position of the HAMR, $\theta_r$ (Figure 4d). The plotted actions are color-coded according to $\Delta\theta_r$ (Figure 4e), with red actions indicating positive forward motion and blue actions indicating retrograde motion. The majority of actions taken by each of the three policies during evaluation resulted in positive motion. Each of the three policies followed similar patterns, in which the phase angle of the X coil was held relatively constant for all $\theta_r$, and the magnitude of the X coil varied according to $\theta_r$. The Y coil was controlled by actuating the phase angle as a function of position and holding the magnitude relatively constant. One consistent pattern across all learned policies is that the magnitudes tended to hold steady close to the maximum or minimum values of -1 and positive 1, regardless of $\theta_r$. This would result in the largest amplitude current sine waves, which we would expect because stronger magnetic fields would be able to create more powerful torques on the HAMR.

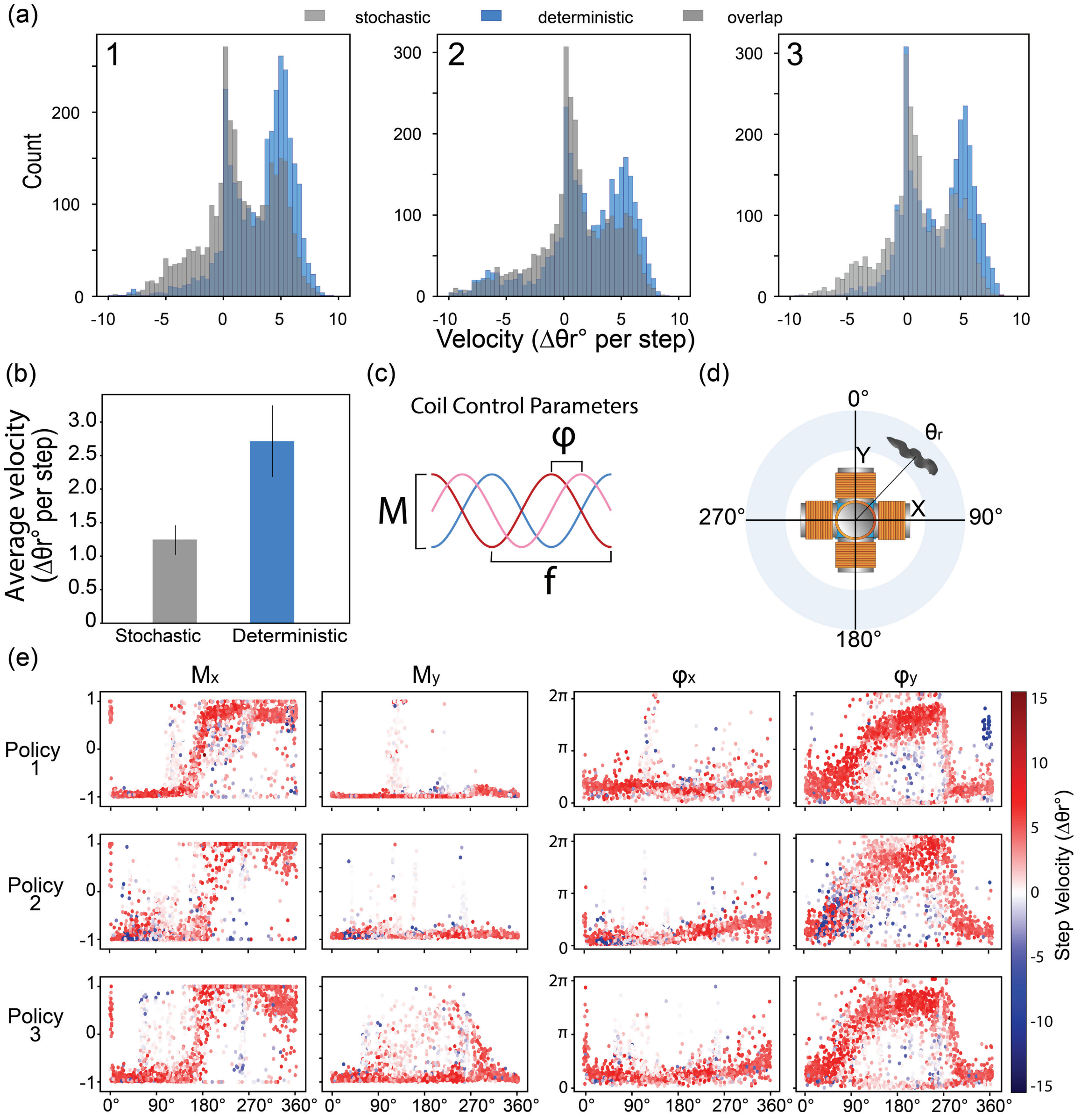

**Figure 4. Evaluating the learning performance of the RL agent during multiple training sessions. (a)** Following three training sessions for the RL agent, we evaluated the performance of the learned stochastic policy, $\pi$, and the deterministic policy $\mu$ for 3k steps without additional learning. In each policy, the performance was higher while selecting actions deterministically from $\mu$. **(b)** The average velocity over 3k evaluation steps for each of the three trained policies during evaluation for stochastic and deterministic action selection, ± standard deviation. **(c)** Schematic showing the control parameters for the sine waves used to drive the current in the Magneturret. The RL agent had control over the Magnitude M and the phase angle $\varphi$ in the X and Y coils. **(d)** Schematic showing the angular position $\theta_r$ of the robot in the circular channel. **(F)** Actions taken by the policies during deterministic evaluation plotted according to $\theta_r$ at the beginning of the action, and color coded according to the resultant velocity during that action.

**Optimizing the RL-trained policies via regression**

We observed that the microrobot control policies learned by the RL agent sometimes performed actions that were obviously non-optimal (resulting in negative motion). We could likely further increase the performance of the learned polices by using techniques like hyperparameter tuning and longer training times [64]. However, by observing the behavior of the RL agent, we hypothesized that if we could distill the policies learned by the network into mathematical functions of the state variables, we might achieve a higher level of performance (Figure 5a). To test this, we chose one of the policies, and fit regression models to the data in order to create continuous control signals as a function of the robot position $\theta_r$. First, we examined policy 1 (Figure 4e). This policy was acting by modulating the magnitude in the X coil in what approximated a square wave pattern and modulating the phase angle in the Y coil in what appeared closer to a sine wave. The other two actions were held approximately constant regardless of the position of the robot. From all 3000 actions taken during policy evaluation, we selected the subset of actions taken by policy which had resulted in a positive movement of at least 3°, discarding the lower performing actions for this analysis. We then fit sinusoidal regression models to the $M_x$ and $\varphi_y$ action distributions, and also fit a square wave to $M_x$ (Figure 5b). The resulting policies are shown in Figure 5a as solid black lines superimposed over the action distribution. The sine wave policy (Figure 5c) and the mixed sine/square wave policy (Figure 5d) that we developed with the regression models were then used to control the HAMR.

The sinusoidal policy achieved the highest level of performance (Figure 5e), achieving the highest average HAMR velocity of all policies tested in this study, while the square/sine policy performed slightly worse than the neural network policy on which it was based. Since these mathematical policies only use $\theta_r$ as the input, it is possible that we could further increase the performance of mathematically inferred policies by taking other parts of the state vector into account.

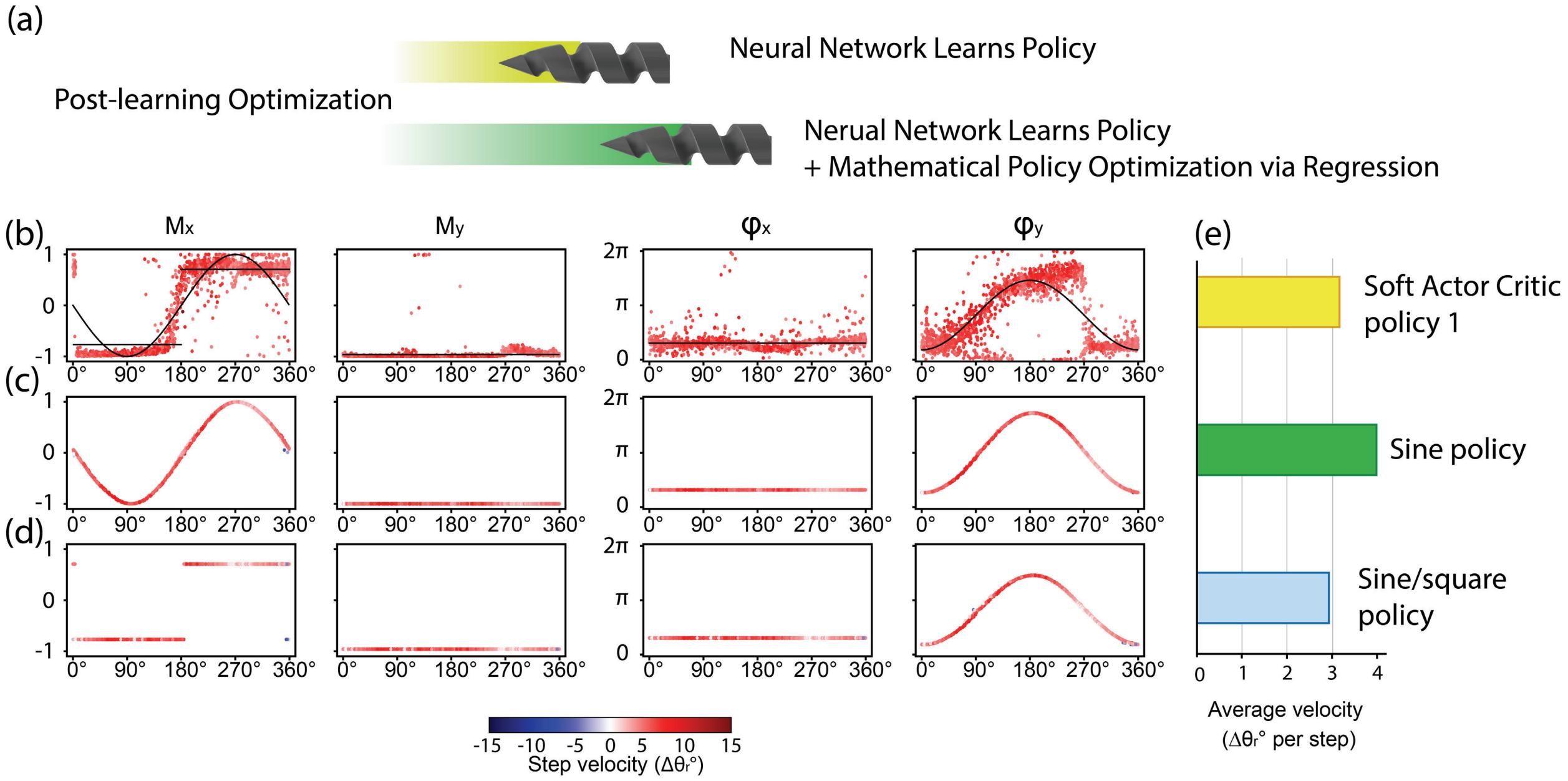

**Figure 5. Control policies learned by the RL agent could be translated into continuous functions in order to increase performance. (a)** The policies learned by the RL agent were used as a basis to derive higher-performance policies via regression of the positive action distribution. **(b)** For one of the policies that the RL agent learned, the actions were plotted as a function of $\theta_r$ and mathematical functions were fit to the subset of actions, which yielded HAMR velocities greater than 3 degrees per step. Sine waves and square waves were fit to the data via regression (shown in black), and those mathematical function were used to control the HAMR for 1k time steps. **(c)** The results of running the sinusoidal policy and **(d)** the sine/square policy. **(e)** Comparing the average velocity of the HAMR when controlled by each policy.

## Discussion

Here, we have reported the development of a closed-loop control system for magnetic helical microrobots, which was implemented using reinforcement learning to discover control policies without the need for any dynamic system modeling. Continuous control policies for high-dimensional action spaces were represented by deep neural networks for effective control of magnetic fields to actuate a helical microrobot within a fluid-filled lumen. Compared with previously reported control systems for magnetic microrobots [23], we believe that the system we have presented possesses a number of key advantages. Electromagnetic actuation systems for microrobots are either air core, such as Helmholtz coils and Maxwell coils, or contain soft magnetic materials in the core which enhance the strength of the generated magnetic field, but can lead to nonlinearities when summing the combined effect of fields from multiple coils with saturated magnetic cores [66]. Such nonlinearities make modeling the behavior of the system more difficult [67], particularly when the coils are run with high enough power to magnetically saturate the core material. Additionally, when controlling microrobots with permanent magnets, those magnets are often modeled as dipole sources for simplicity [68], and the actual behavior of the physical system may not match the idealized model behavior. Neural network-based controllers trained with RL learn control policies from observing the actual behavior of the physical system, and deep neural networks can accurately model non-linear functions [69]. Control policies learned with RL will automatically take into account the real system dynamics, and this model-free control approach can greatly simplify the job of the microrobotic engineer.

Significant work has been done to create accurate dynamic models of rigid helical microrobots, and many microrobots are relatively straightforward to control using these models and classical control systems [47]. However, many recently developed microrobotic systems are composed of soft, shape changing materials, which are inherently harder to model than rigid bodies [5, 13, 14, 16, 17]. Soft microrobots such as helical grippers, which can kinematically change their configuration during operation, might also be difficult to accurately model [7]. Here, we have shown that our algorithm was able to control a soft helical microrobot without any dynamic modeling on the part of the control system designers. RL based microrobot control could enhance both the capabilities of novel microrobot designs and increase the efficiency of researchers by allowing the RL agent to do the work of developing a high-performance controller. RL based controllers may be able to exceed the performance of classical control systems based on simplified models (e.g., linearized models) because the RL agent is able to learn based on the observed physical behavior of the system, and deep neural networks are capable of accurately modeling any observed nonlinearities that the microrobotic system might exhibit.

Other control strategies that have been successfully applied for microrobot control could be enhanced with RL. Algorithms which have been used to control soft microrobots such as force-current mapping with PID control and path planning algorithms [5, 23, 67] could potentially be combined with reinforcement learning in order to optimize the gains in the PID controllers, and adapt to changes in environmental conditions by a process of continuous learning, or to optimize for multiple variables. Force-current mapping algorithms used to control microrobots are also often created with assumptions of linearity in magnetic field superposition, which could be violated with soft magnetic cores in the driving coils [50], a limitation that could be potentially overcome by the nonlinear function approximation capabilities of deep neural networks. Another potential combination of RL with classical control has been demonstrated by Zeng et al [39], who used a residual physics model in order to fine tune a physics-based model of a grasping and throwing problem for a robotic arm. Using such methods, imprecise models of microrobot dynamics could be improved by fine tuning their parameters based on a data driven RL approach, leading to increased performance.

We have demonstrated that it is possible to learn microrobot control policies with reinforcement learning based on no prior knowledge, and then fine tune the performance of the policy by fitting continuous mathematical functions to the learned policy behaviors. While the sine and sine/square policies that we created based on analyzing the learned policies might have been identified by a first principles analysis of the problem, the policies derived by the neural networks could potentially uncover useful behaviors, which would not be suspected or created by human engineers, particularly for systems that are difficult to model. This ability to detect subtle patterns from high dimensional data in a model-free RL approach could ultimately lead to state-of-the-art control policies that exceed the performance of human-designed policies, as has been seen with reinforcement learning algorithms in tasks like playing Go [41] and classic Atari games [63].

In our experimentation, we found that the RL agent could learn successful polices from state vector inputs, but this system could likely be expanded to use and raw image data as input, as has been demonstrated in other RL-based robotic control research [56]. The ability to derive control models directly from high-dimensional data could make RL-based microrobot control applicable for a broad class of biomedical imaging modes in which the state of the system might be represented by MRI, X-ray, ultrasound, or other biomedical imaging methods [24]. Using higher dimension inputs like images has the potential to encode richer policies which respond to objects in the field of view such as obstacles which could impede the forward progress of the microrobot but would not be observable from lower dimensional feedback available in a state vector representation. In complex environments in which environmental factors such as lumen shape, fluid flow profiles, surface interactions, and biological interactions are likely to be a significant factor [2], the ability to use machine vision for state representation could significantly improve microrobot performance. All these points strongly favor the use of RL for developing the next generation of microrobot control systems.

## Methods

**Helical agar magnetic robot.** The HAMR was constructed based on a method published by Hunter et al [4]. The structure of the robot was formed from a 2% w/v agar-based hydrogel (Fisher Cat. No. BP1423-500). The agar was melted to above 80 degrees Celsius and mixed with iron oxide nanopowder (Sigma Aldrich Cat. No. 637106) to a total concentration of 10% w/v. This mix was injected into a helical 3D printed mold printed on the Elegoo Mars stereolithography 3D printer to form a helical microrobot 4.4 mm in length. The microrobot was manually removed from the mold after cooling and solidifying, and stored in deionized water until use. Because the yield of this batch fabrication technique was not 100%, robots used for subsequent experiments were chosen based on their morphology and responsiveness to magnetic fields.

**Circular swimming arena.** The PDMS swimming arena was created by molding Sylgard 184 elastomer (Sigma Aldrich Cat. No. 761036) over a thin 3mm tall section of polyvinyl chloride pipe (31 mm Inner Diameter, 34mm Outer diameter) Access holes for the microrobot were cut, and then the molded PDMS was plasma bonded to a thin uniform sheet of PDMS to close the channel, and cured overnight at 65° C.

**Magneturret.** The Magneturret was constructed by winding 6 identical coils with 400 turns each of 30-gauge magnet wire (Remington Industries Cat. No. 30H200P) around a 0.26 inch diameter permalloy core (National Electronic Alloys Cat. No. HY-MU 80 Rod .260 AS DRAWN) cut to a length of 20 mm. These coils were fixed to the sides of an Acrylonitrile butadiene styrene (ABS) 3D printed cube with quick set epoxy. The coil was enclosed in a 3D printed housing printed in Zortrax Z-glass filament with a Zortrax M200 printer and sealed with epoxy. Glycerol coolant was pumped through the housing with a liquid CPU cooling system (Thermaltake Cat. No. CL-W253-CU12SW-A). The coils were energized by creating sinusoidal currents with an Arduino STEMtera breadboard, which took serial commands from the RL agent over USB and turned them into PWM signals which were sent to two Pololu Dual G2 High-Power Motor Driver 24v14 Shields. The power supply used to power the coils and was a Madewell 24V DC power supply.

**Overhead camera**. The overhead camera was an Alvium 1800 U-500c with a 6mm fixed focal length lens from Edmund Optics. The camera used to take images for the state was set at a long exposure so that the HAMR and the center mark were the only visible objects in the image. A second identical camera placed above the arena at a slight angle was used to simultaneously record normal exposure video of the HAMR in the arena during operation, so that the features in the image were not washed out.

**RL algorithm**. The soft actor critic RL agent was developed in Python, using Tensorflow 2.0 for creating the neural network models. This was run on a desktop workstation from Lambda Labs. Separate processes were used for data collection and updating the neural networks so that the two operations could run in parallel. The full algorithm details are available in Supplementary Algorithm 1.

**Supplementary Materials:**

Fig S1: Highest performing policy parameters during training

Table S1: Hyperparameters

Movie S1: Training process time course.

Movie S2: HAMR swimming around the track

## Acknowledgements

**Funding:**

National Institutes of Health through the Director's New Innovator Award, DP2-GM132934 (WCR)

National Science Foundation, DMR 1709238 (WCR)

Air Force Office of Scientific Research, FA9550-18-1-0262 (WCR)

Office of Naval Research, N00014-17-1-2306 (WCR)

National Institutes of Health Cellular Approaches to Tissue Engineering and Regeneration Training Program traineeship T32- EB001026 (MRB)

William Kepler Whiteford Faculty Fellowship from the Swanson School of Engineering at the University of Pittsburgh (WCR)

**Author Contributions:**

Conceptualization: MRB and WCR

Methodology: MRB and WCR

Investigation: MRB

Visualization: MRB

Funding Acquisition: WCR and MRB

Writing –original draft: MRB and WCR

Writing – Review and Editing: MRB and WCR

**Competing Interests:**

The authors declare that they have no competing interests.

**Acknowledgments:**

We gratefully thank Haley C. Fuller and Ting-Yen Wei for scientific discussions and manuscript suggestions.

**Data and Materials Availability:**

All data are available in the main text or the supplementary materials. The code used in this work is available at https://github.com/Synthetic-Automated-Systems/RUDER_MBOT_RL

## SUPPLEMENTARY MATERIALS

# Smart Magnetic Microrobots Learn to Swim with Deep Reinforcement Learning

Michael R. Behrens[1] and Warren C. Ruder[1,2,*]

[1]Department of Bioengineering
University of Pittsburgh, Pittsburgh, PA, USA

[2]Department of Mechanical Engineering
Carnegie Mellon University, Pittsburgh, PA, USA

*Corresponding author: warrenr@pitt.edu

## Contents

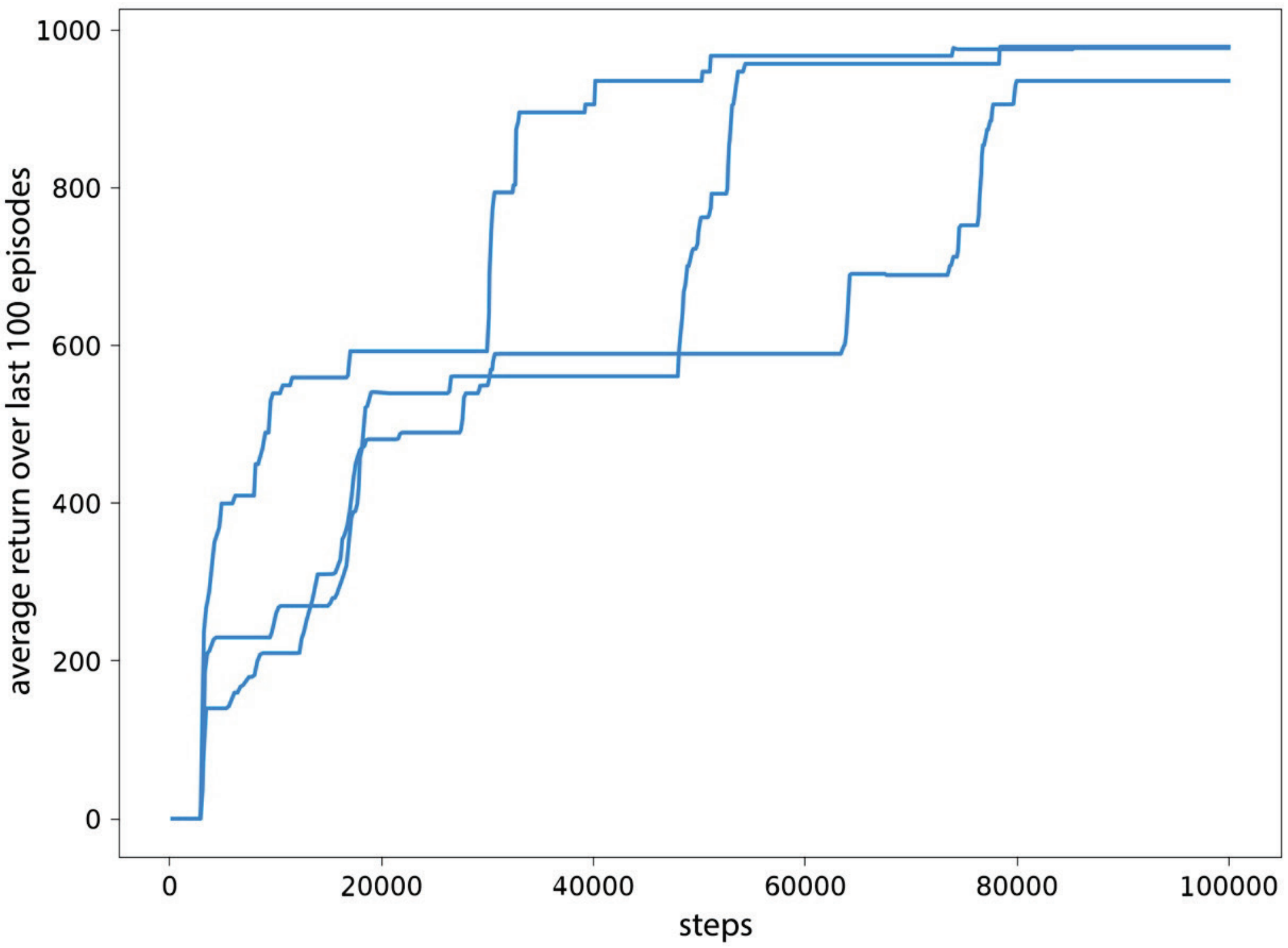

**Supplementary Figure S1. Selecting the highest performing policy parameters during training.** During training of the three policies, the average return from the policy was continually monitored with a rolling average of the last 100 episodes. Whenever the average return during the last 100 episodes was higher than at any point previously recorded during training, the parameters (i.e., the weights and biases) of the actor network, $\pi$, were saved. For evaluating the performance of the policy after learning, the highest performing parameters saved during the training session were used, rather than the policy parameters of the network at the end of the training session. This was done because we sometimes observed a decrease in the performance of the policy at the end of training.

**Supplementary Table 1: Hyperparameters**

| | |
|---|---|
| State actor network architecture | 1. Input Shape: (7,3)<br>2. (1)→Dense layer with 256 neurons. activation: Relu<br>3. (2)→Dropout (0.2)<br>4. (3)→Dense layer with 256 neurons. Activation: Relu<br>5. (4)→Dropout(0.2)<br>6. (5)→Output shape: (4) |
| State critic network architecture | 1. State Input Shape: (7,3)<br>2. (1)→Dense layer with 16 neurons, activation: Relu<br>3. Action Input Shape: (4)<br>4. (3)→Dense layer with 16 neurons, activation: Relu<br>5. (2,4)→Concatenate Sate and Action Input<br>6. (5)→Dense layer with 256 neurons. activation: Relu<br>7. (6)→Dropout (0.2)<br>8. (7)→Dense layer with 256 neurons. Activation: Relu<br>9. (8)→Dropout(0.2)<br>10. (9)→Output shape: (1) |
| Convolutional actor network architecture | 1. Input Shape: (64,64,3)<br>2. (1)→2D convolutional layer, 16 filters, 3x3 kernel, activation: Relu<br>3. (2)→Max pooling(2,2)<br>4. (3)→2D convolutional layer, 32 filters, 3x3 kernel, activation: Relu<br>5. (4)→Max pooling(2,2)<br>6. (5)→2D convolutional layer, 64 filters, 3x3 kernel, activation: Relu<br>7. (6)→Max pooling(2,2)<br>8. (7)→Flatten<br>9. (8)→Dense layer with 64 neurons. activation: Relu<br>10. (9)→Dense layer with 256 neurons. activation: Relu<br>11. (10)→Dropout (0.2)<br>12. (11)→Dense layer with 256 neurons. Activation: Relu<br>13. (12)→Dropout(0.2)<br>14. (13)→Output shape: (4) |
| Convolutional critic network architecture | 1. State Input Shape: (64,64,3)<br>2. (1)→2D convolutional layer, 16 filters, 3x3 kernel, activation: Relu<br>3. (2)→Max pooling(2,2)<br>4. (3)→2D convolutional layer, 32 filters, 3x3 kernel, activation: Relu<br>5. (4)→Max pooling(2,2)<br>6. (5)→2D convolutional layer, 64 filters, 3x3 kernel, activation: Relu<br>7. (6)→Max pooling(2,2)<br>8. (7)→Flatten<br>9. (8)→Dense layer with 64 neurons. activation: Relu<br>10. Action Input Shape: (4)<br>11. (10)→Dense layer with 16 neurons, activation: Relu<br>12. (9,11)→Concatenate Sate and Action Inputs<br>13. (12)→Dense layer with 256 neurons. activation: Relu<br>14. (13)→Dropout (0.2)<br>15. (14)→Dense layer with 256 neurons. Activation: Relu<br>16. (15)→Dropout(0.2)<br>17. (16) →Output Shape: (1) |
| Experience replay buffer size: | 100,000 |
| Batch Size | 256 |
| Learning Rate | 0.0003 |
| Gamma | 0.99 |
| Tau | 0.005 |
| Number of Actions | 4 |
| Target Entropy | -4 (negative dim of number of actions) |
| Policy Update Period | 1 minute |
| Total Step Duration | 0.9 Seconds (three concatenated observations spaced 0.3 seconds apart) |
| Goal Distance | 20 Degrees |
| Episode Length | 33 Steps |
| Gradient update steps per environmental step: | 1 |

**Supplementary movies and code are available on Github at:**

https://github.com/Synthetic-Automated-Systems/RUDER_MBOT_RL

**Supplementary Movie S1: Training process time course.** This movie shows an accelerated time course of the HAMR's movements during training. The agent starts out randomly exploring actions, and as training progresses the HAMR begins to move continuously in a clockwise direction around the arena.

**Supplementary Movie S2: HAMR swimming around the track.** This movie shows the swimming action of the HAMR in real time, for one complete trip around the circular track during evaluation of state-based policy No. 1.